\documentclass[conference]{IEEEtran}
\IEEEoverridecommandlockouts
\usepackage{cite}
\usepackage{amsmath,amssymb,amsfonts}
\usepackage{algorithmic}
\usepackage{graphicx}
\usepackage{textcomp}
\usepackage{xcolor}
\usepackage{hyperref}
\def\BibTeX{{\rm B\kern-.05em{\sc i\kern-.025em b}\kern-.08em
    T\kern-.1667em\lower.7ex\hbox{E}\kern-.125emX}}
\begin{document}

\title{Efficient Label Refinement for Face Parsing Under Extreme Poses Using 3D Gaussian Splatting\\
}


\author{
Ankit Gahlawat, Anirban Mukherjee, Dinesh Babu Jayagopi\\
\textit{International Institute of Information Technology, Bangalore (IIIT-B)}\\
Bengaluru, India\\
\{ankit.gahlawat002, anirban.mukherjee, jdinesh\}@iiitb.ac.in
}

\maketitle

\IEEEpubid{%
  \makebox[\columnwidth]{}%
  \hspace{\columnsep}%
  \makebox[\columnwidth]{%
    \parbox{\columnwidth}{\centering\scriptsize
    \textcopyright~2025 IEEE. Personal use of this material is permitted.
    Permission from IEEE must be obtained for all other uses, including
    reprinting/republishing for advertising or promotional purposes,
    creating new collective works, for resale or redistribution to servers or lists,
    or reuse of any copyrighted component of this work in other works.}%
  }%
}

\begin{figure*}[t]
    \centering
    \includegraphics[width=\linewidth]{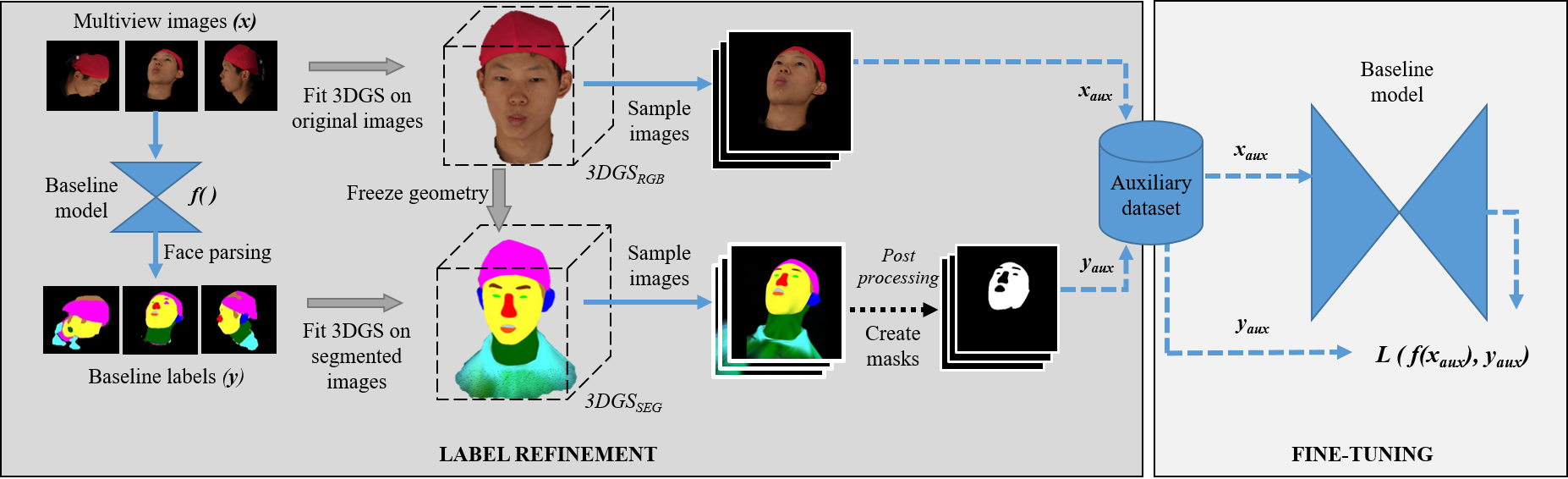}
    \caption{Overview of our two-stage label refinement pipeline. We use multiview images and coarse baseline predictions to fit dual 3DGS models with shared geometry, enabling multiview-consistent synthesis of dense segmentation renderings. These are clustered and manually refined to produce training labels used for fine-tuning the baseline model.}

    \label{fig:pipeline}
    \vspace{-0.5cm}
\end{figure*}

\begin{abstract}
Accurate face parsing under extreme viewing angles remains a significant challenge due to limited labeled data in such poses. Manual annotation is costly and often impractical at scale. We propose a novel label refinement pipeline that leverages 3D Gaussian Splatting (3DGS) to generate accurate segmentation masks from noisy multiview predictions. By jointly fitting two 3DGS models, one to RGB images and one to their initial segmentation maps, our method enforces multiview consistency through shared geometry, enabling the synthesis of pose-diverse training data with only minimal post-processing. Fine-tuning a face parsing model on this refined dataset significantly improves accuracy on challenging head poses, while maintaining strong performance on standard views. Extensive experiments, including human evaluations, demonstrate that our approach achieves superior results compared to state-of-the-art methods, despite requiring no ground-truth 3D annotations and using only a small set of initial images. Our method offers a scalable and effective solution for improving face parsing robustness in real-world settings.
\end{abstract}


\section{Introduction}

Face parsing, the task of semantically segmenting facial regions such as eyes, lips, and nose, is essential for a wide range of vision applications including facial recognition, animation, and augmented reality \cite{khan2020face}. While state-of-the-art methods~\cite{yu2018bisenet,xie2021segformer, narayan2024facexformer} perform well under frontal or near-frontal views, they often fail to generalize to extreme head poses due to the strong frontal-view bias in existing annotated datasets.

Popular benchmarks like CelebAMask-HQ~\cite{CelebAMask-HQ}, Helen~\cite{le2012interactive}, and LaPa~\cite{liu2020new} provide detailed pixel-wise annotations, but are limited in pose diversity. As a result, face parsing models trained on these datasets tend to exhibit poor performance in real-world conditions involving profile or top-down views.

Previous approaches have attempted to address this limitation by synthesizing novel viewpoints using 3D face models~\cite{lin2021roi,rawat2024synthforge}. However, these methods predominantly generate limited in-plane variations, such as left–right rotations, and often under represent more challenging out-of-plane poses, including top-down views or combinations of side and elevated angles. In our setup, we take care to include such extreme viewpoints to ensure broader pose coverage during training.

In this work, we propose a label refinement pipeline that leverages \textit{3D Gaussian Splatting} (3DGS)~\cite{kerbl20233d}. 3DGS represents a scene using a set of anisotropic 3D Gaussian primitives, each defined by its center $\mu$, covariance matrix $\Sigma$, opacity $\alpha$, and spherical harmonic color coefficients. The contribution of a Gaussian to a pixel $x$ is modeled as:
\[
G(x) = \alpha \cdot \exp\left(-\frac{1}{2}(x - \mu)^T \Sigma^{-1} (x - \mu)\right),
\]
and the final color is rendered by alpha-blending depth-sorted Gaussians:
\[
C = \sum_{i \in N} c_i \alpha_i \prod_{j=1}^{i-1} (1 - \alpha_j).
\]
This formulation is differentiable, spatially smooth, and robust to sparse, noisy observations, making it ideal for label refinement across views.

Given a set of RGB images and coarse segmentation maps from a baseline model, we fit two aligned 3DGS models: one for appearance and one for masks, sharing a common geometry. This shared geometry acts as a consistency prior across viewpoints and enables multiview aggregation to reduce label noise. Rendering these models from novel viewpoints yields paired RGB and segmentation images across diverse poses, which are refined with minimal manual cleanup to create an auxiliary dataset $(x_{\text{aux}}, y_{\text{aux}})$ for fine-tuning face parsing networks under extreme poses.





We validate our pipeline by fine-tuning BiSeNet \cite{yu2018bisenet} on this refined dataset. Despite using only 77 training images across 6 identities, the fine-tuned model demonstrates significantly improved robustness on both held-out identities and out-of-distribution scenes, achieving performance competitive with state-of-the-art methods in human evaluations.

\noindent \textbf{Our main contributions are:}
\begin{itemize}\IEEEpubidadjcol
    \item We propose a novel label refinement pipeline based on dual 3D Gaussian Splatting (3DGS), using shared geometry and multiview consistency to generate accurate, pose-diverse labels without 3D ground truth.
    
    \item We demonstrate that fine-tuning a face parsing model on these minimally refined labels yields strong improvements under extreme poses.
    
    \item Despite using limited training data, our method achieves competitive or superior performance compared to recent state-of-the-art models in human evaluations.

\end{itemize}

\vspace{0.5cm}

\section{Proposed Method}

Our method consists of two stages: (1) a 3DGS-based label refinement pipeline that aggregates multiview predictions into high-quality renderings, and (2) fine-tuning a parsing model using these improved labels. Fig.~\ref{fig:pipeline} illustrates the overall workflow.

\noindent \textbf{1. Initial Face Parsing.} 
We begin with multi-view RGB images from FaceScape ~\cite{yang2020facescape, zhu2023facescape}, using a baseline model ~\cite{yu2018bisenet} to generate initial, often noisy and inconsistent, segmentation maps, especially under challenging poses.

\noindent \textbf{2. 3DGS Fitting.}  
We fit two 3DGS models: one to the RGB images (\(3DGS_{RGB}\)) and one to the segmented images (\(3DGS_{SEG}\)), using the frozen geometry of \(3DGS_{RGB}\) for both. During fitting, the 3DGS model mitigates noise by aggregating consistent features across views, producing a coherent 3D representation of facial segments. Since the segmented images lack geometric cues like shading and depth, transferring the geometry from \(3DGS_{RGB}\) enables accurate reconstruction. This shared geometry is also critical for rendering spatially aligned image-label pairs used later for fine-tuning.

\noindent \textbf{3. Viewpoint Sampling.}  
We render both 3DGS models from identical virtual viewpoints, simulating pose diversity. The rendered RGB images constitute the inputs for our auxiliary dataset and are denoted by \(x_{aux}\). The segmentation renderings are further processed to derive the corresponding labels \(y_{aux}\).

\noindent \textbf{4. Label Clustering and Post-processing.}  
Rendered segmentation images may exhibit lighting-induced color gradients. To recover discrete masks, we apply k-d tree clustering~\cite{bentley1975multidimensional} to group similar colors corresponding to each semantic region. These masks are then minimally refined through manual editing to improve accuracy. This primarily involves correcting small artifacts and occasionally, refining semantics of fine features like eyes or lips. The resulting outputs form the labels \(y_{aux}\) for our fine-tuning step.

\noindent \textbf{5. Model Fine-tuning.}  
The refined segmentation masks \(y_{aux}\) and their corresponding RGB inputs \(x_{aux}\) are used to fine-tune the baseline model. This improves its generalization to diverse and extreme head poses with minimal manual labeling effort.

\section{Experiments}

\subsection{Experimental setup}

\noindent \textbf{Dataset:} To construct the 3D Gaussian Splatting (3DGS) models, we utilize a subset of the FaceScape~\cite{zhu2023facescape, yang2020facescape} multi-view dataset, which contains RGB images captured from diverse viewing angles. Our subset includes eight identities (four male, four female), each with distinct facial expressions to introduce expression-level variability. Of these, six identities (77 images) are used to generate the auxiliary dataset, while the remaining two identities (15 images) are held out for evaluation. For sampling of the images, we utilized the SuperSplat\footnote{https://playcanvas.com/supersplat/editor} editor. Both $3DGS_{RGB}$ and $3DGS_{SEG}$ were sampled from the viewpoints mentioned in Table ~\ref{tab:sampling_angles}.


\begin{table}[htbp]
\caption{Sampling angles (in degrees) used for $3DGS_{RGB}$ and $3DGS_{SEG}$, organized by axis. Each column represents one sampled camera pose}
\centering
\scriptsize  
\setlength{\tabcolsep}{1.5pt}  
\renewcommand{\arraystretch}{1.1}
\begin{tabular}{|l|*{19}{c}|}
\hline
\textbf{Axis} & 1 & 2 & 3 & 4 & 5 & 6 & 7 & 8 & 9 & 10 & 11 & 12 & 13 & 14 & 15 & 16 & 17 & 18 & 19 \\
\hline
\textbf{x} & 0 & 0 & 180 & 180 & 45 & 45 & 35 & 30 & 40 & 30 & -170 & -150 & -45 & -65 & -45 & -45 & 135 & 135 & 0 \\
\textbf{y} & 45 & 25 & -65 & -45 & 40 & 10 & -10 & -30 & -45 & -70 & -65 & -50 & 30 & 5 & -30 & -90 & -65 & -40 & -45 \\
\textbf{z} & 0 & 0 & 180 & 180 & 0 & 0 & 0 & 0 & 0 & 0 & 180 & 180 & 0 & 0 & 0 & 0 & 180 & 180 & 0 \\
\hline
\end{tabular}
\label{tab:sampling_angles}
\end{table}

\noindent \textbf{Baseline Model:} For all experiments, we use the BiSeNet model~\cite{yu2018bisenet}, trained on the CelebAMask-HQ dataset~\cite{CelebAMask-HQ}, as our baseline. BiSeNet is widely adopted in face parsing for its real-time efficiency and lightweight design, with public implementations readily available. However, as shown in \cite{kvanchiani2023easyportrait}, its segmentation accuracy is notably lower than recent state-of-the-art models, particularly in fine-grained facial regions. This choice is deliberate: due to the absence of labeled datasets for extreme head poses, there is no established benchmark for evaluating parsing models in such scenarios. By using a lower-performing baseline, we establish a controlled setup to assess the benefit of our refined supervision. If even a lightweight model like BiSeNet can match or exceed state-of-the-art performance after fine-tuning with our generated labels, it highlights the effectiveness and generalizability of our pipeline. While our experiments are based on BiSeNet, the proposed label refinement method is model-agnostic and applicable to other face parsing architectures. We focus on seven commonly used facial classes: face, eyebrows, eyes, ears, nose, lips, and neck. 

\noindent \textbf{Evaluation metrics:} To evaluate our approach, we use mean Intersection over Union (mIoU) and F1 score. These are the most commonly used metrics for segmentation tasks. The mIoU measures the overlap between predicted segmentation maps and the ground truth, while the F1 score balances precision and recall to assess the accuracy of the segmented regions.

\begin{table}
\caption{Comparison of baseline and 3DGS-based parsing results across facial regions}
\centering
\resizebox{1\linewidth}{!}{
    \begin{tabular}{|l||c|c||c|c|}
    \hline
    
    \textbf{Labels} & \textbf{Baseline mIoU $\uparrow$} & \textbf{3DGS mIoU $\uparrow$} & \textbf{Baseline F1 Score $\uparrow$} & \textbf{3DGS F1 Score $\uparrow$} \\
    \hline
    \textbf{Face}   & 0.46 & 0.98 & 0.59 & 0.99 \\
    \textbf{Eyebrows} & 0.14 & 0.41 & 0.19 & 0.52 \\
    \textbf{Eyes}   & 0.06 & 0.31 & 0.08 & 0.36 \\
    \textbf{Ears}   & 0.23 & 0.97 & 0.27 & 0.98 \\
    \textbf{Nose}   & 0.20 & 0.99 & 0.24 & 0.99 \\
    \textbf{Lips}   & 0.06 & 0.39 & 0.07 & 0.51 \\
    \textbf{Neck}   & 0.26 & 0.43 & 0.32 & 0.60 \\
    \hline
    \end{tabular}
}
\label{tab:quantitative_results1}
\vspace{-0.3cm}
\end{table}

\begin{figure}[t]
    \centering


    \begin{minipage}{0.22\linewidth}
        \includegraphics[width=\linewidth]{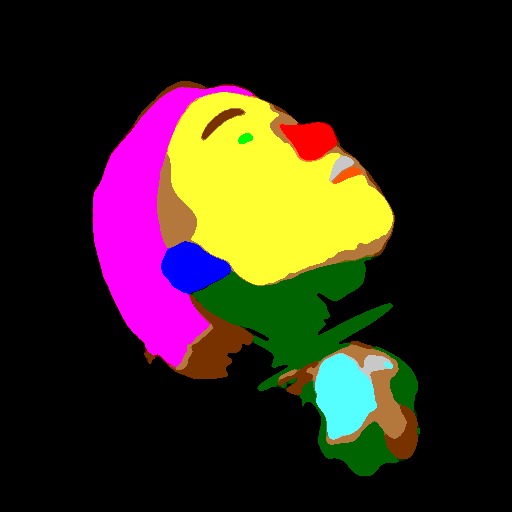}
    \end{minipage}
    \hspace{-0.02\linewidth}
    \begin{minipage}{0.22\linewidth}
        \includegraphics[width=\linewidth]{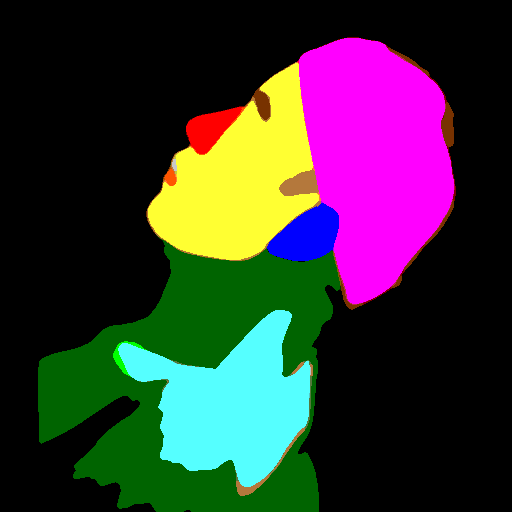}
    \end{minipage}
    \hspace{-0.02\linewidth}
    \begin{minipage}{0.22\linewidth}
        \includegraphics[width=\linewidth]{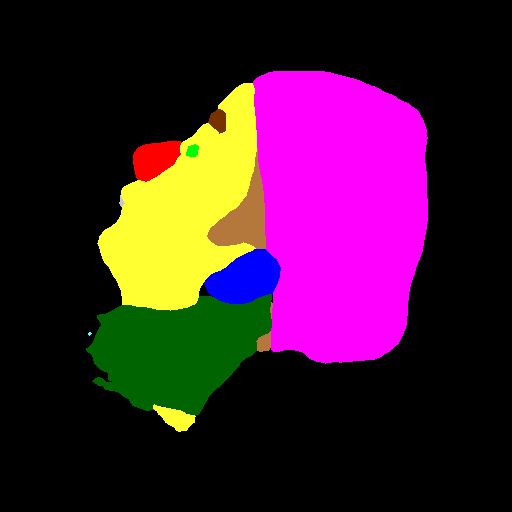}
    \end{minipage}
    \hspace{-0.02\linewidth}
    \begin{minipage}{0.22\linewidth}
        \includegraphics[width=\linewidth]{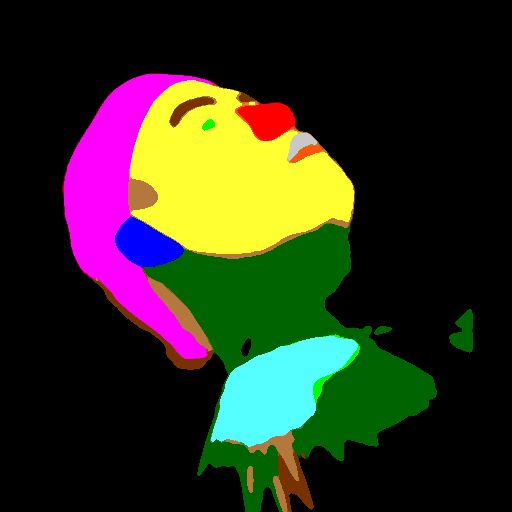}
    \end{minipage}

    \vspace{1pt}
    \begin{minipage}{0.22\linewidth}
        \includegraphics[width=\linewidth]{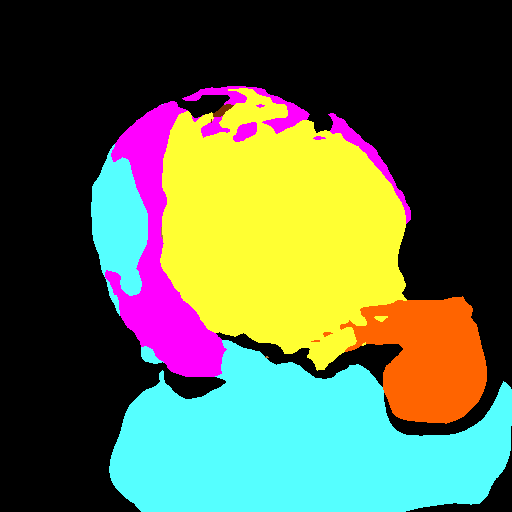}
    \end{minipage}
    \hspace{-0.02\linewidth}
    \begin{minipage}{0.22\linewidth}
        \includegraphics[width=\linewidth]{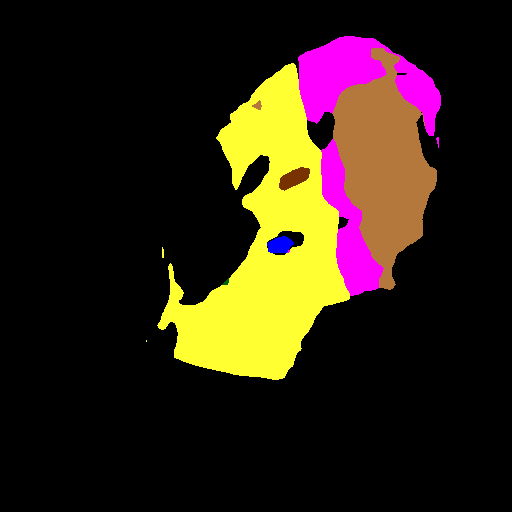}
    \end{minipage}
    \hspace{-0.02\linewidth}
    \begin{minipage}{0.22\linewidth}
        \includegraphics[width=\linewidth]{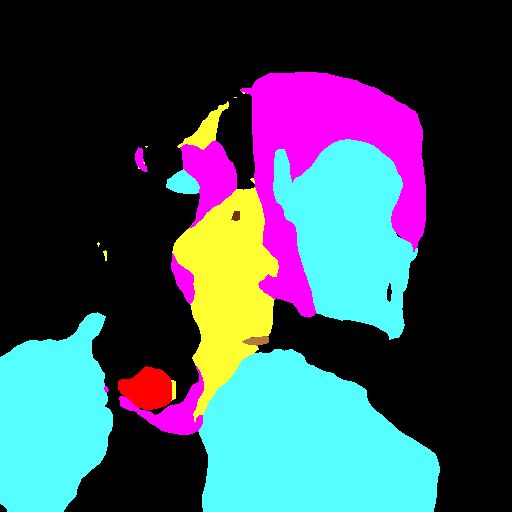}
    \end{minipage}
    \hspace{-0.02\linewidth}
    \begin{minipage}{0.22\linewidth}
        \includegraphics[width=\linewidth]{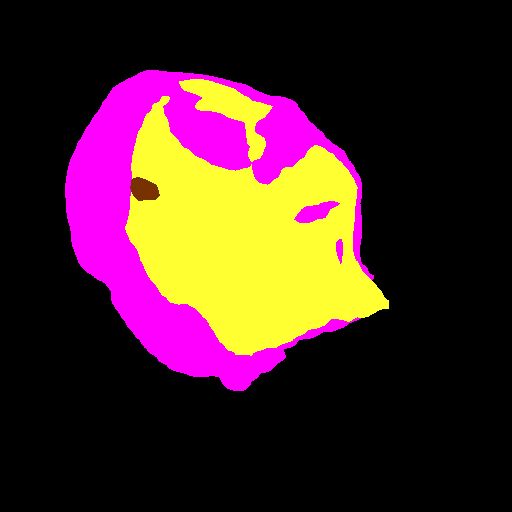}
    \end{minipage}

    \vspace{1pt}

    \begin{minipage}{0.22\linewidth}
        \includegraphics[width=\linewidth]{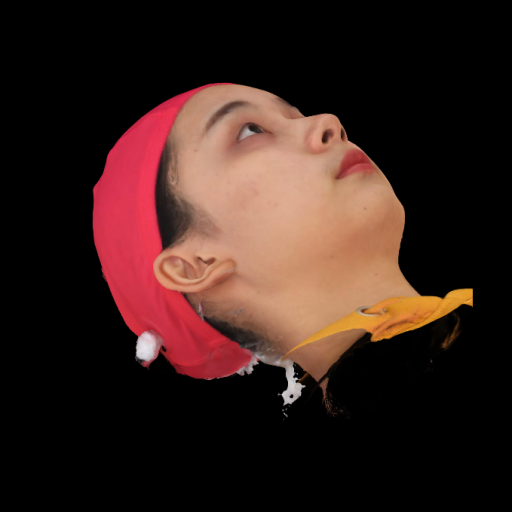}
    \end{minipage}
    \hspace{-0.02\linewidth}
    \begin{minipage}{0.22\linewidth}
        \includegraphics[width=\linewidth]{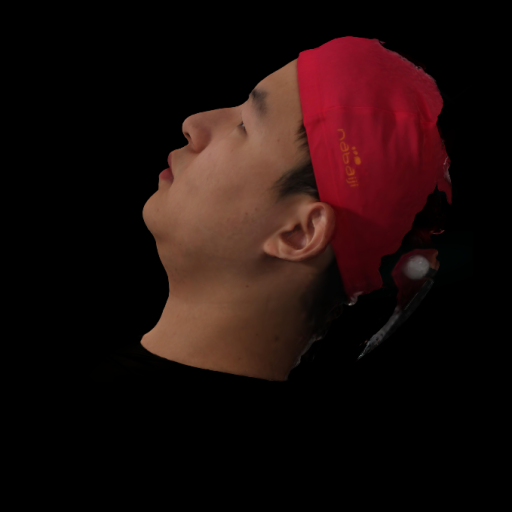}
    \end{minipage}
    \hspace{-0.02\linewidth}
    \begin{minipage}{0.22\linewidth}
        \includegraphics[width=\linewidth]{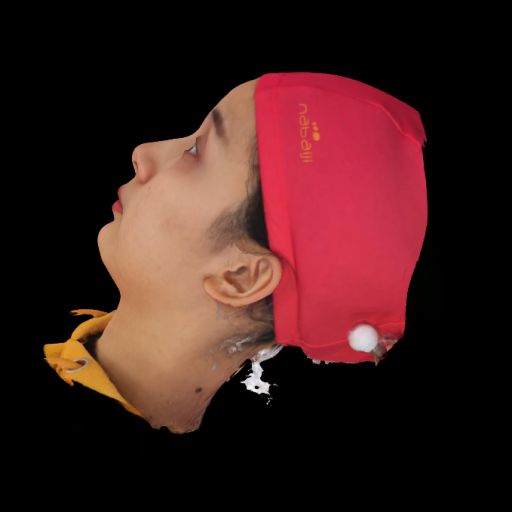}
    \end{minipage}
    \hspace{-0.02\linewidth}
    \begin{minipage}{0.22\linewidth}
        \includegraphics[width=\linewidth]{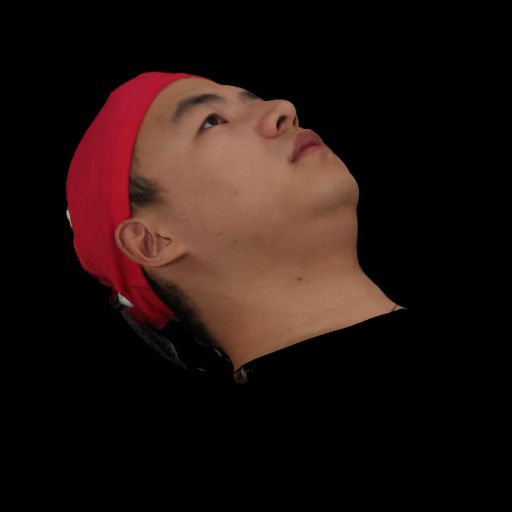}
    \end{minipage}

    \vspace{1pt}
    \caption{Face parsing on extreme view images \textit{(bottom)} using the baseline model \textit{(middle)} and our automatic 3DGS-based refinement \textit{(top)}. Our method produces cleaner, more consistent segmentations without manual supervision.}
    \label{fig:baseline_vs_3DGS_extended}
    \vspace{-0.5cm}
\end{figure}

\begin{table}[b]
\caption{Table III. Performance on held-out test identities before and after fine-tuning}
\centering
\resizebox{1\linewidth}{!}{
\begin{tabular}{|c||c|c||c|c|}
\hline
\textbf{Ids} & \textbf{Baseline mIoU $\uparrow$} & \textbf{Finetuned mIoU $\uparrow$} & \textbf{Baseline F1 Score $\uparrow$} & \textbf{Finetuned F1 Score $\uparrow$} \\
\hline
\textbf{id\_1} & 0.32 & 0.53 & 0.39 & 0.66 \\
\hline
\textbf{id\_2} & 0.27 & 0.68 & 0.33 & 0.77 \\
\hline
\end{tabular}
}
\label{tab:miou_f1_comparison}
\vspace{-0.5cm}
\end{table}

\begin{figure}[b]
    \vspace{-0.5cm}
    \centering
    \includegraphics[width=0.45\textwidth]{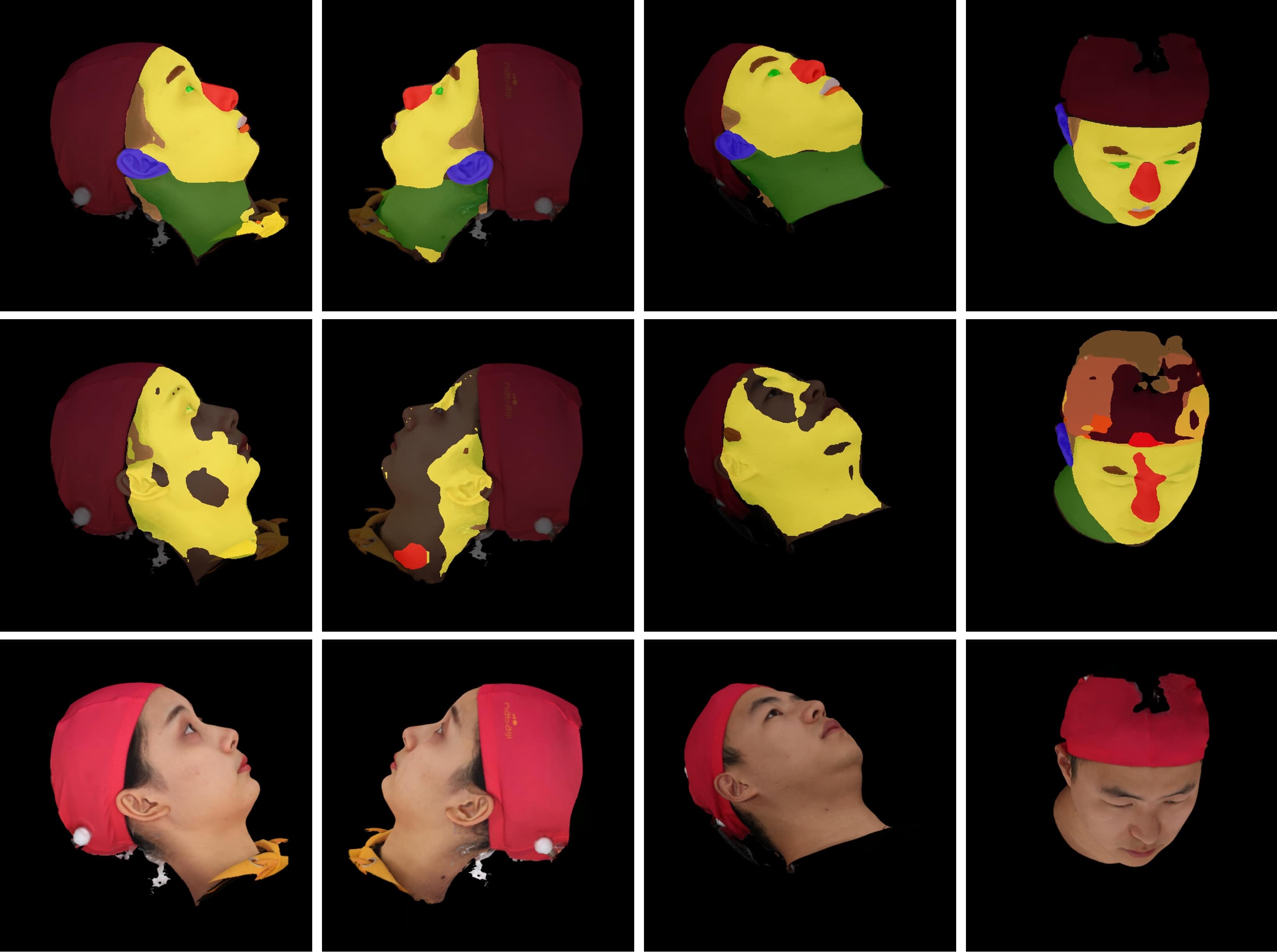}
    \caption{Face parsing on held-out test images \textit{(bottom)} using our baseline model before \textit{(middle)} and after fine-tuning on the auxiliary dataset using our method \textit{(top)}.}
    \label{fig:results_test}
    \vspace{-0.5cm}
\end{figure}

\subsection{Results}

We organize our experiments into two stages corresponding to the main components of our pipeline: (1) Label Refinement, where we leverage 3D Gaussian Splatting to enhance noisy segmentation maps by enforcing multiview consistency without requiring manual 3D supervision; and (2) Fine-Tuning, where the refined labels, after minimal manual post-processing, are used to improve the baseline face parsing model’s performance under extreme pose variations. We evaluate each stage through standard segmentation metrics and perceptual analysis, including a user study.
\noindent \subsubsection{\textbf{Label Refinement via 3DGS}}

Our first set of experiments evaluates the quality of segmentation masks obtained after 3D Gaussian Splatting (3DGS) fitting. By aggregating information from multiple views, 3DGS effectively suppresses noise and inconsistencies present in initial baseline segmentations. As shown in Table~\ref{tab:quantitative_results1}, the refined masks exhibit substantial improvements in both mIoU and F1 score across all key facial regions, with especially large gains in difficult areas such as the ears, nose, and lips.

This performance gain is achieved without any additional manual annotation, demonstrating the scalability and practicality of our method in minimizing human labeling effort. Qualitative comparisons in Fig.~\ref{fig:baseline_vs_3DGS_extended} further support this, showing that 3DGS masks are visually cleaner, more complete, and closely aligned with ground truth, particularly under challenging head poses.

\noindent \subsubsection{\textbf{Fine-Tuning with Refined Labels}}
Following minimal manual corrections to the segmentation masks generated by 3DGS, we utilize the refined labels to fine-tune the baseline face parsing model and assess its generalization across both held-out identities and out-of-distribution settings.

\noindent \textbf{Held-Out Test Set.}
We evaluate the fine-tuned model on two identities held out from the fine-tuning dataset. As shown in Table~\ref{tab:miou_f1_comparison} , we observe substantial improvements in both mIoU and F1 score, aggregated over all facial regions. Qualitative comparisons in Fig.~\ref{fig:results_test} reveal clear visual enhancements over the baseline model, particularly in delineating challenging regions such as ears and facial contours under extreme poses. These results indicate that supervision derived from our 3DGS-based pipeline can meaningfully improve face parsing performance, even with a small number of labeled identities.

\begin{figure}[htbp]
    \centering
    \includegraphics[width=0.45\textwidth]{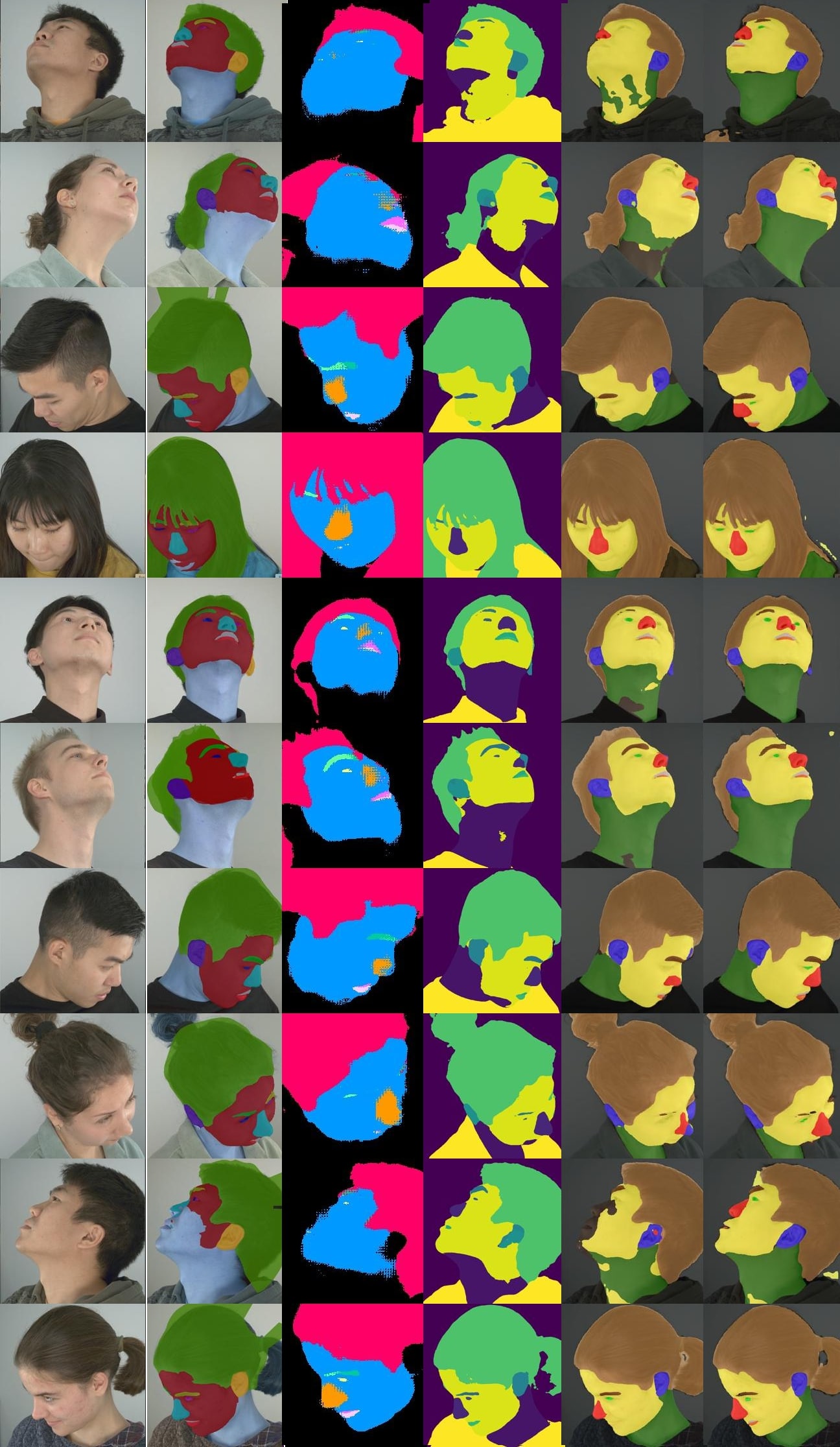}
    \caption{Face parsing results on out-of-distribution images \textit{(leftmost column)} using various models \textit{(columns left to right)}: 
    RoI Tanh-polar Transformer~\cite{lin2021roi}, FaceXFormer~\cite{narayan2024facexformer}, SegFormer~\cite{xie2021segformer}, 
    BiSeNet~\cite{yu2018bisenet} (baseline), and BiSeNet fine-tuned with our 3DGS-based auxiliary dataset. Despite using only 77 images, our method generalizes well to novel subjects and poses using minimally refined labels.}
    \label{fig:ood}
    \vspace{-0.5cm}
\end{figure}

\noindent \textbf{Out-of-Distribution Evaluation.}
To further assess generalization, we test our fine-tuned model on the NeRSemble \cite{kirschstein2023nersemble} dataset: an entirely unseen multi-view face dataset captured under conditions distinct from our training set. While no ground-truth labels are available for this dataset, we perform a human user study to quantitatively assess perceptual quality and generalization. We additionally provide qualitative results that demonstrate strong performance on novel subjects and extreme viewpoints (see Fig.~\ref{fig:ood})).

The user study involved 24 participants. Each participant was shown a set of 10 images, sampled from a total of 40 images grouped into four distinct sets, and asked to rate segmentations from five different state-of-the-art models, including our fine-tuned BiSeNet, on a 1–5 Likert scale based on visual accuracy and completeness. As shown in Table~\ref{tab:mean_scores}, our model - despite being fine-tuned on just 77 images, achieves the highest perceptual scores among all models, confirming its effectiveness and generalization to challenging real-world scenarios.

\begin{table}[t]
\centering
\caption{User study results comparing state-of-the-art models}
\resizebox{1\linewidth}{!}{
\begin{tabular}{|l||c|c|c|}

\hline
\textbf{Model} & \textbf{Mean Score} & \textbf{Std. Dev} & \textbf{Median Score} \\

\hline
ROI-TanH \cite{lin2021roi} & 3.41 & 0.21 & 3.5 \\
FaceXFormer \cite{narayan2024facexformer} & 1.52 & 0.42 & 1.0  \\
SegFormer \cite{xie2021segformer} & 3.37 & 0.28 & 3.0  \\
Baseline \cite{yu2018bisenet} & 3.45 & 0.21 & 3.5  \\
\textbf{Baseline-Finetuned \textit{(Ours)}} & \textbf{4.22 }& \textbf{0.19 }& \textbf{4.25 } \\
\hline

\end{tabular}
}
\label{tab:mean_scores}
\vspace{-0.5cm}
\end{table}

\begin{figure}[b]
    \centering
    \includegraphics[width=0.45\textwidth]{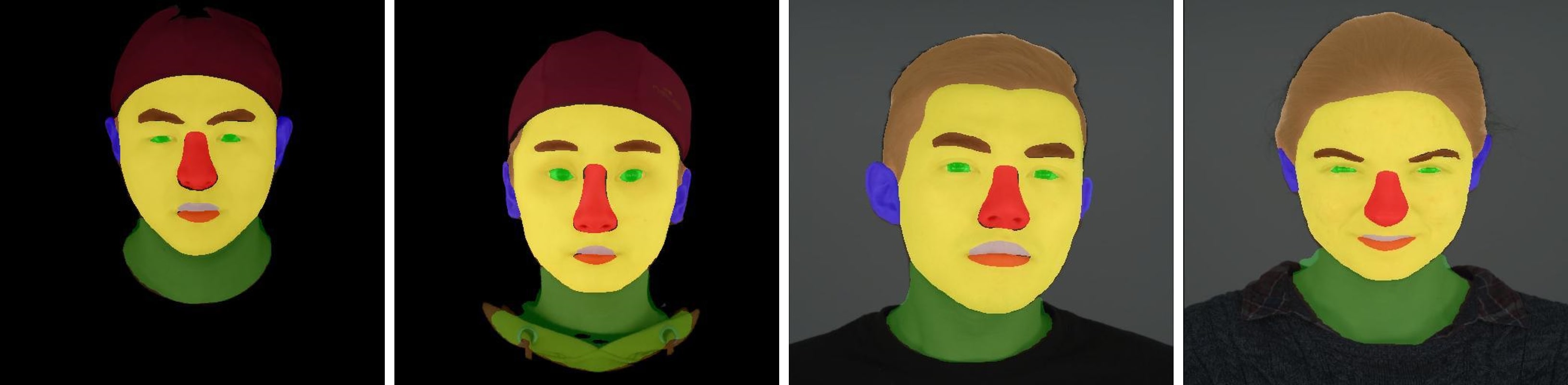}
    \caption{Face parsing results on frontal views, showing that our fine-tuned model maintains strong performance on standard poses.}
    \label{fig:front_poses}
    \vspace{-0.5cm}
\end{figure}

\noindent \textbf{Pose-General Robustness.}
Importantly, as shown in Fig.~\ref{fig:front_poses}, our fine-tuned model retains high accuracy on frontal face images, confirming that improvements on extreme poses do not come at the cost of performance on standard views. This underscores the effectiveness of our refined supervision in boosting model robustness across diverse head poses without introducing regressions on the original task.



\section{Discussion}

Our approach demonstrates that accurate, pose-diverse segmentation masks can be synthesized from a small set of noisy predictions by leveraging the multiview consistency and geometry-aware rendering properties of 3D Gaussian Splatting. While we use only six identities in this work, the quality of the refined masks and their impact on fine-tuning suggest that even limited supervision can be bootstrapped into effective label generation. Increasing identity and expression diversity would likely improve generalization to a broader population and more varied facial configurations. Furthermore, as refinement quality is bounded by initial predictions, starting with a stronger baseline model could significantly amplify gains. This opens an exciting direction: iteratively reapplying our refinement pipeline with additional data, treating the fine-tuned model as a new baseline, to progressively enhance segmentation quality without manual 3D ground truth.

\section{Conclusion}
In this work, we present a novel method to enhance the robustness and accuracy of face-parsing models by utilizing 3D Gaussian Splatting to generate an auxiliary dataset for fine-tuning. Our method synthesizes highly precise segmentation masks from noisy predictions by learning a 3D head model, especially under extreme or unseen viewing angles. With minimal manual intervention, these masks significantly improve baseline model performance. Despite using only 77 training images, our method generalizes well to new identities and poses. Furthermore, our label refinement pipeline is model-agnostic and can yield greater gains when applied to stronger face parsing architectures. Overall, the approach reduces labeling effort and improves pose robustness, making it practical for real-world applications.

\section*{Acknowledgment}
This project was funded by MINRO center at IIIT Bangalore.

\bibliographystyle{acm}
\bibliography{refs}

\end{document}